\documentclass{article}

\usepackage{PRIMEarxiv}
\usepackage{amsmath}
\usepackage[utf8]{inputenc} 
\usepackage[T1]{fontenc}    
\usepackage{hyperref}       
\usepackage{url}            
\usepackage{booktabs}       
\usepackage{amsfonts}       
\usepackage{nicefrac}       
\usepackage{microtype}      
\usepackage{lipsum}
\usepackage{fancyhdr}       
\usepackage{graphicx}       
\graphicspath{{media/}}     

\title{Learning the spatio-temporal relationship between wind and significant wave height using deep learning} 

\author{
  Said Obakrim\\
  IRMAR \\
  Université de Rennes 1\\
  \texttt{said.obakrim@univ-rennes1.fr} \\
   \And
  Valérie Monbet \\
  IRMAR \\
  Université de Rennes 1\\
  \texttt{valerie.monbet@univ-rennes1.fr} 
  \And
  Nicolas Raillard\\
  LCSM\\
  Ifremer\\
  \texttt{nicolas.raillard@ifremer.fr} 
  \And
  Pierre Ailliot\\
  LMBA\\
  Université de Bretagne Occidentale\\
  \texttt{pierre.ailliot@univ-brest.fr} 
}

\begin{document}

\maketitle


\begin{abstract}
    Ocean wave climate has a significant impact on near-shore and off-shore human activities, and its characterisation can help in the design of ocean structures such as wave energy converters and sea dikes. Therefore, engineers need long time series of ocean wave parameters. Numerical models are a valuable source of ocean wave data; however, they are computationally expensive. Consequently, statistical and data-driven approaches have gained increasing interest in recent decades. This work investigates the spatio-temporal relationship between North Atlantic wind and significant wave height ($H_s$) at an off-shore location in the Bay of Biscay, using a two-stage deep learning model. The first step uses convolutional neural networks (CNNs) to extract the spatial features that contribute to $H_s$. Then, long short-term memory (LSTM) is used to learn the long-term temporal dependencies between wind and waves.
\end{abstract}

\section{Introduction}

Characterisation of wave climate is required for many marine applications, such as the design of coastal and offshore structures and the planning of ship operations. Wind waves are generated by the surface wind, with local wind creating the wind sea and wind from distant areas creating waves that propagate and form swells (\cite{waves}). Waves in the Bay of Biscay depend on both local and large-scale wind conditions in the North Atlantic (\cite{charles}) ; however, swells generally dominate the sea state. Swells travel large distances and take up to five days to cross the Atlantic from Cape Hatteras to the Bay of Biscay (\cite{inbook}). Consequently, waves observed at a given location depend on wind conditions over the North Atlantic in a time window of several days, and it is challenging to reproduce this complex spatio-temporal relationship using machine learning. The goal of this work is to propose a deep learning approach that learns this relationship.

 The advantage of deep learning methods (\cite{goodf}) lies in their ability to build hierarchical representations of predictors. In particular, in the case of spatial data, convolutional neural networks (CNNs) allow to learn complex spatial features from the data (\cite{cnn}). Moreover, long short memory (LSTMs) (\cite{lstm}) have proven to be very successful in predicting time series and sequence data. In this work, we propose a non-expensive data-driven approach that learns the underlying spatio-temporal structure of the relationship between wind and waves using a two-stage model based on CNNs and LSTM.

This paper is organised as follows. Section 2 presents the problem of downscaling of ocean waves and related works. Section 3 describes the data used in this work. Section 4 presents the proposed two-stage model, the architecture, and the training process. Section 5 discusses the results of this work. Finally, Section 6 presents the conclusions and future work directions. 

\section{Problem statement and related work}
The problem of improving the spatial resolution of climate variables is known under the name of downscaling (\cite{maraun}). Downscaling approaches attempt to construct a link, either numerical or statistical, between large-scale and local-scale variables. The advantage of statistical downscaling (SD) over numerical models is primarily in terms of computational efficiency. A rigorous comparison of the two approaches can be found in (\cite{wang, laugel}).  

In the case of ocean waves, wind (\cite{paper}) or sea level pressure (SLP) (\cite{camus}) are commonly used to downscale ocean wave parameters. However, in order to establish a link function between the wind (or SLP) and the local ocean wave parameters, it is necessary to consider a large spatial and temporal coverage and, consequently, a large number of potential explanatory variables that are highly correlated. Some methods determine the wave generation area for any ocean location worldwide. For example, ESTELA (\cite{estela}) is a numerical model that uses the spectral information to select the fraction of energy that travels to the target point from selected source points. The ESTELA method can be used to design statistical downscaling methods. For instance, \cite{camus} and \cite{heg} used the ESTELA method to define the predictors used in their SD model.  

\cite{paper} proposed a data-driven approach that determines the wave generation area by estimating the travel time of waves, generated in each considered sources point, that reach the target point.
Then, the predictors were defined based on the wave generation area and finally a SD model based on weather types was built.  

As far as we know, the existing methods for SD of ocean wave parameters define a priori the spatio-temporal structure of the predictors, and then the SD model is built using these predictors. The aim of this study is to propose a deep learning approach that automatically learns the spatio-temporal relationship between wind and waves.  

\section{Data preparation}

The Climate Forecast System Reanalysis (CFSR) (\cite{cfsr}) hourly wind data is considered in this study as a predictor. CFSR is a global reanalysis developed by the National Centers for Environmental Prediction (NCEP) that covers the period from 1979 to the present with an hourly time step and a spatial resolution of 0.5\textdegree  by 0.5\textdegree.
The historical $H_s$ data is extracted from the hindcast database HOMERE (\cite{homere}) at the target location with spatial coordinates (45.2\textdegree N, 1.6\textdegree W) located in the Bay of Biscay. The temporal resolution of both wind and $H_s$ data is up-scaled to 3-hourly data. The period from 1994 to 2016 is considered in this study, leading to a dataset with $n = 67208$ observations. 

\begin{figure}
\centering
\includegraphics[width=90mm]{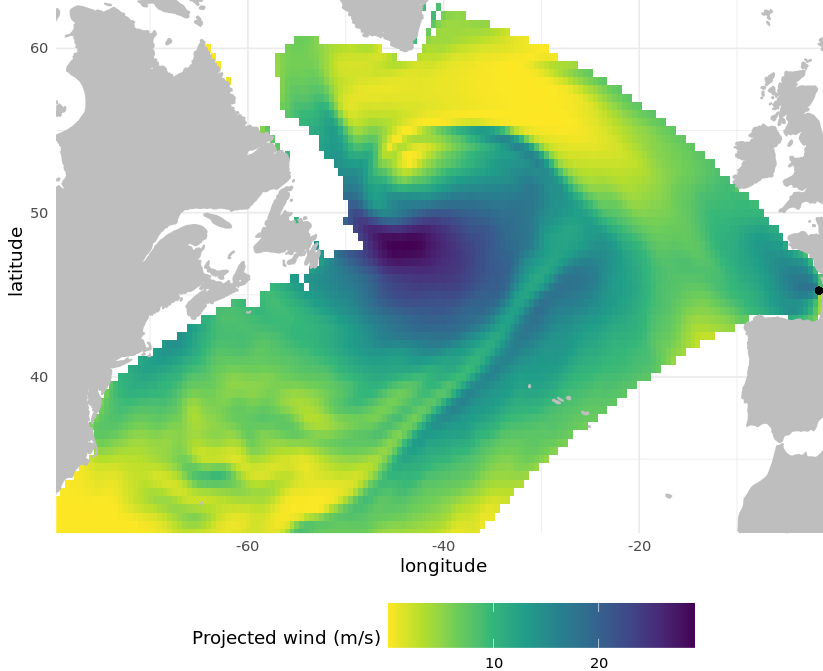}
\caption{The projected wind defined in \eqref{eq:eq2} in 1994-01-01 00h:00. The black point represents the target point}
\label{fig:fig1}
\end{figure}

Instead of using both zonal and meridional components as a predictor, we use the projected wind (\cite{paper}) defined, at each location $j$ and time $t$, as
\begin{equation}
W_j(t) = U_j(t)\,cos^{2}\left(\frac{1}{2}(b_j-\theta_j(t))\right)\\
\label{eq:eq1}
\end{equation}
where $W_j(t)$ is the projected wind, $U_j(t)$ the wind speed, $\theta_j(t)$ the wind direction and $b_j$ is the great circle bearing from the source point $j$ to the target point. Under the assumption that waves travel in great circle paths, grid points whose paths are blocked by land are neglected (Figure \ref{fig:fig1}). 
Therefore, we define the global predictor at time $t$ as
\begin{equation}
    X^{(g)}(t) = (W_1^2(t),..., W_p^2(t))
    \label{eq:eq2}
\end{equation}
where $p = 5651$ is total number of grid points.

Following (\cite{paper}), in order to capture the wind sea, we also define the local predictor as 
\begin{equation}
        X^{(\ell)}(t) = \{U(t),U^2(t),U^3(t),U^2(t)F(t),U(t-1), U^2(t-1),U^3(t-1),U^2(t-1)F(t-1)\}
\label{eq:eq3}
\end{equation}
where $U(t)$ is the wind speed at the target point and $F(t)$ is the fetch length at time $t$, calculated as the minimum of the distance from the target point to shore in the direction from which the wind is blowing and  $500km$. The fetch has an important effect on wind sea characteristics (\cite{inbook}); therefore, it is commonly used to construct empirical wind wave models.  

\section{Proposed methodology}

As mentioned in the last section, state of the art statistical methods for downscaling wave parameters usually use a pre-processing step to create features that take into account the wave generation area. In this study, we propose a deep learning approach that automatically extracts these features. Since waves may take several days to reach the target point, the history and current wind can be used to predict $H_s$. An example of this type of model could have the following form
\begin{equation}
    H_s(t) = f(X^{(g)}(t-t_{max}),...,X^{(g)}(t))
    \label{eq:eq4}
\end{equation}
where, $t_{max}$ can be interpreted as the maximum travel time of the waves and will be referred to as such in the following. However, this approach can be computationally challenging given the dimension of the predictor (5651 in our case). Instead, in this study we propose to use current wind conditions to estimate current and future $H_s$. 

In order to describe the complex spatio-temporal relationship between wind and $H_s$, we propose the following two-stage model

\begin{equation}
   \begin{split}
     \textbf{1\textsuperscript{st} stage:}\,\,\,&  [H_s(t|X^{(g)}(t)),...,H_s(t+t_{max}|X^{(g)}(t))] = f(X^{(g)}(t)) + \epsilon(t),\,\,\, f:\mathbb{R}^p\rightarrow \mathbb{R}^{t_{max}} \\ 
     \textbf{2\textsuperscript{nd} stage:}\,\,\, &H_s(t) = g(X^{(g)}(t),f(X^{(g)}(t-t_{max})),...,f(X^{(g)}(t))) + \epsilon\prime(t),\,\,\, g:\mathbb{R}^{t_{max}*t_{max}+8}\rightarrow \mathbb{R} 
    \end{split}
    \label{eq:eq5}
\end{equation}
where the notation $H_s(t_1|X^{(g)}(t_2))$ represents the  contribution of wind conditions at time $t_2$ in $H_s$ at time $t_1$. $\epsilon$ and $\epsilon \prime$ are the errors of the \text{1\textsuperscript{st} stage} and \text{2\textsuperscript{nd} stage}, respectively. 
The \text{1\textsuperscript{st} stage} estimates the current and future $H_s$ using  current wind conditions. The \text{2\textsuperscript{nd} stage} estimates $H_s$ using the past predictions obtained from the \text{1\textsuperscript{st} stage}. Along with the local predictor $X^{(g)}$, the input for the \text{2\textsuperscript{nd} stage} is a $t_{max}*t_{max}$ matrix of the form
\begin{equation}
\begin{pmatrix}
\hat{H}_s(t-t_{max}|X^{(g)}(t-t_{max}))&\dots& \hat{H}_s(t|X^{(g)}(t-t_{max}))\\
\vdots & \ddots & \vdots \\
\hat{H}_s(t|X^{(g)}(t))&\dots& \hat{H}_s(t+t_{max}|X^{(g)}(t))
\end{pmatrix}
\label{eq:eq6}
\end{equation}
where $\hat{H}_s(t_1|X^{(g)}(t_2))$ represents the prediction, obtained from the \text{1\textsuperscript{st} stage}, of the contribution of wind conditions at time $t_2$ in the $H_s$ at time $t_1$. When $t_1 = t_2$, this prediction represents the wind sea (first column of the matrix in equation \eqref{eq:eq6}); for $t_1 > t_2$, on the other hand, the prediction represents the $H_s$ caused by swells.
\begin{figure}
\centering
\includegraphics[width=145mm]{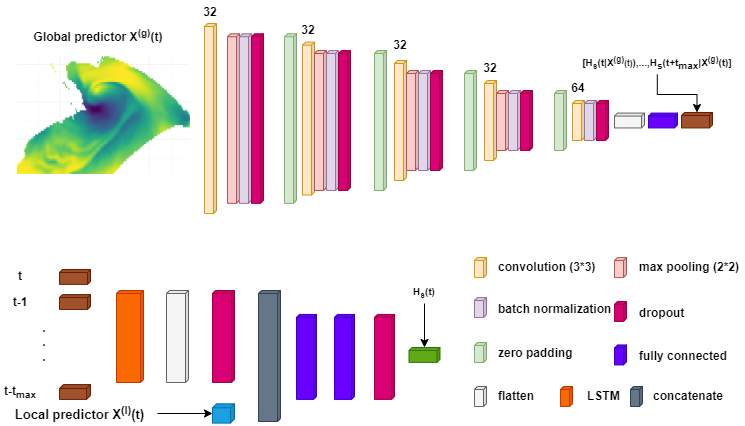}
\caption{Architecture of the two-stage model in equation \eqref{eq:eq5}}
\label{fig:fig2}
\end{figure}

 The general structure of the model is shown in Figure \ref{fig:fig2}. The \text{1\textsuperscript{st} stage} consists of a series of 3*3 convolutions followed by the ReLU activation function, 2*2 max pooling layer, Batch Normalisation, then a flatten followed by a dense layer. The \text{2\textsuperscript{nd} stage} starts with an LSTM layer that learns the long-term dependencies of the $(t-t_{max},...,t)$ outputs of the \text{1\textsuperscript{st} stage}. The output of the LSTM layer is then concatenated with the local predictor $X^l$ and fed into two fully connected layers. The dropout layer is used in both stages to prevent the network from overfitting. The loss function choosed in this study is the mean squared error (MSE) which is expressed as
\begin{equation}
\begin{split}
    MSE(\text{1\textsuperscript{st} stage}) &= \frac{1}{t_{max}} \sum_{i=0}^{t_{max}} \frac{1}{n-t_{max}-1} \sum_{t=1}^{n-t_{max}} (H_s(t+i) - \hat{H}_s(t+i|X^{(g)}(t)))^2\\
    MSE(\text{2\textsuperscript{nd} stage}) &= \frac{1}{n} \sum_{t=1}^{n} (H_s(t) - \hat{H}_s(t))^2
\end{split}
\label{eq:eq7}
\end{equation} 
Where $n$ is the total number of observations and $\hat{H}_s$ is the prediction of $H_s$. The Keras framework with Tensorflow backend (\cite{keras}) is used in this work to train the model, on a Nvidia K80s GPU using the Adam optimiser (\cite{adam}) and mini batches of 64.
\section{Results}
\begin{figure}
\centering
\includegraphics[width=110mm]{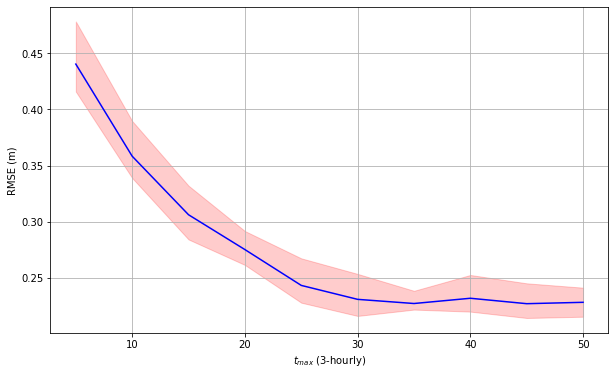}
\caption{Results of cross-validation using different values of $t_{max}$. The blue line represents the mean of RMSE and the red interval represents the minimum and maximum RMSE}
\label{fig:fig3}
\end{figure}
The period from 1994 to 2011 is used to train the two-stage model and the period from 2012 to 2014 serves as the validation period. The measures chosen in this paper to validate the analysis are the correlation coefficient (r), the root mean square error (RMSE) and the bias. Different values for the maximum travel time of waves $t_{max}$ are tested, and the results of k-fold cross-validation (with $k=5$) are shown in Figure \ref{fig:fig3}. The RMSE stabilises approximately at $t_{max} = 30\times3h$, which corresponds to about 3.3 days, and the gain is substantial compared to using $t_{max} = 5$. This means that wind conditions over a time window of at least 3.3 days must be considered to characterise the wave climate at the target location. In the following, the value of $t_{max}$ is chosen equal to $30$.

\begin{figure}
\centering
\includegraphics[width=150mm]{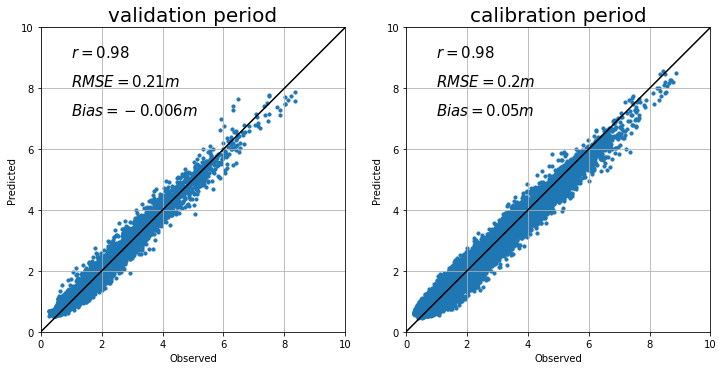}
\caption{Observed versus predicted $H_s$ in the validation period (left panel) and calibration period (right panel)}
\label{fig:fig4}
\end{figure}
\begin{figure}
\centering
\includegraphics[width=110mm]{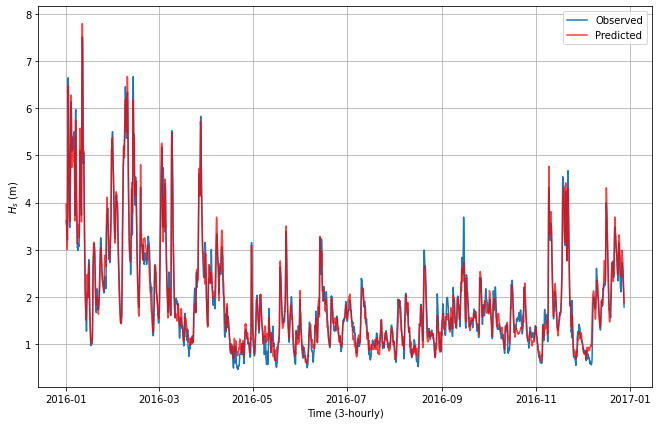}
\caption{Time series of observed (blue line) and predicted (red line) $H_s$ in 2016}
\label{fig:fig5}
\end{figure}

Figure \ref{fig:fig4} shows the scatter plot of observed versus predicted values of $H_s$ using the two-stage model \eqref{eq:eq5}. The RMSE in the validation period is equal to $0.21m$ for an $H_s$ of mean $1.9m$ and standard deviation $1.1m$. The model performs well in predicting $H_s$ and accounts for both wind and swell. The validation measures in the calibration and validation periods are almost the same. This means that the model does not overfit the training data and generalises well the relationship between wind and waves. Furthermore, the seasonality of $H_s$ is well captured by the two-stage model, as can be seen in Figure \ref{fig:fig5}.

\begin{table}
    \centering
    \begin{tabular}{c c c c}
    \hline
    Method & r & RMSE(m)  & bias(m) \\ \hline
    two-stage model & 0.98 & 0.21 &-0.006\\ 
    weather types  & 0.97 & 0.27 &-0.03\\ 
    H-CNN & 0.97 & 0.27 & -0.04 \\ \hline
    \end{tabular}
    \caption{Comparison of the two-stage model, weather types and H-CNN methods.}
    \label{tab:tab1}
\end{table}

A comparison of the two-stage model with two other statistical approaches is done in Table \ref{tab:tab1}.
The first approach, described in (\cite{paper}), is based on weather types (\cite{maraun}). As for the present work, the local and global predictors were considered.  However, in order to reduce the dimension of the predictor a single predictor is extracted at each spatial location $j$ to predict $H_s$ at time $t$. It is defined a priori as
\begin{eqnarray}
\label{eq:eq8}
   &X^{(g)}_j(t;t_j,\alpha_j) = \frac{1}{2\alpha_j+1} \sum_{i = t-t_j-\alpha_j}^{t-t_j+\alpha_j} W_j^2(i),\\ \nonumber
    &t_j + \alpha_j  +1 \leq t \leq t_j -\alpha_j + n
\end{eqnarray}
where $t_j$ is the travel time of waves, $\alpha_j$ controls the length of the time window, and $W_j$ is the projected wind at location $j$. The parameters $t_j$ and $\alpha_j$ were estimated using the maximum correlation between $h_s$ and the global predictor. The second method (\cite{hcnn}) uses CNNs to predict $H_s$ using the same predictors as in (\cite{paper}). Thus, the main difference with the approach proposed in this work is that the temporal dimension of the global predictor is reduced a priori using the preprocessing step based on maximum correlation described above. The numerical results in Table \ref{tab:tab1} indicate that the two-stage model significantly outperforms the other two methods in term of the validation measures. 

\section{Conclusion}
In this study, a two-stage model based on deep learning is proposed to predict $H_s$ using wind conditions. The model is capable of learning automatically the underlying spatio-temporal structure of the relationship between wind and waves. The model does well in predicting $H_s$ and is computationally inexpensive (about 5min using a computer of 30GB RAM, 2 cores CPU, and a 16GB GPU). The proposed methodology is based on two stages which are trained separately. A natural question that arises for future work, is whether we can estimate the parameters jointly using back-propagation and eventually speed up the training process and improving the results. Future work also includes using the method to predict other sea state parameters, such as wave direction and period.  

The proposed method can be used for climate and weather studies at any ocean location worldwide. For nearby locations, one can train only the \text{2\textsuperscript{nd} stage} at each location, using the weights of one location as initialisation for the others and leaving the \text{1\textsuperscript{st} stage} the same.  
The model can also learn from buoy data instead of hindcast data and eventually fill in the gaps and complete historical data.

\paragraph{Data Availability Statement}
The hindcast data Homere is available in their website: \url{https://marc.ifremer.fr/produits/rejeu_d_etats_de_mer_homere}. The wind data is available from the CFSR website: \url{https://climatedataguide.ucar.edu/climate-data/climate-forecast-system-reanalysis-cfsr}.   

\paragraph{Author Contributions}
Conceptualization: S.O; V.M; N.R; P.A. Methodology: S.O; V.M; N.R; P.A. Data curation: S.O; N.R. Data visualisation: S.O. Software: S.O. Supervision: V.M; N.R; P.A. Writing original draft: S.O; V.M; N.R; P.A. All authors approved the final submitted draft.

\paragraph{Supplementary Material}
For reasons of reproducibility, Python code and the processed data are available at \url{https://github.com/SaidObakrim/Two-stage-CNN-LSTM-}.


\begin{thebibliography}{}

\bibitem[Ardhuin et al. (2018)]{inbook}
\textbf{Ardhuin, Fabrice and Orfila, Alejandro} (2018) Wind Waves, \textit{New Frontiers in Operational Oceanography.} \textit{393--422}

\bibitem[Boudi{\`e}re et al.(2013)]{homere}
\textbf{Boudi{\`e}re, Edwige and Maisondieu, Christophe and Ardhuin, Fabrice and Accensi, Micka{\"e}l and Pineau-Guillou, Lucia and Lepesqueur, J{\'e}r{\'e}my} (2013) A suitable metocean hindcast database for the design of Marine energy converters,  \textit{International Journal of Marine Energy.} \textit{3},  {e40}--{e52}.

\bibitem[Camus et al.(2014)]{camus}
\textbf{Camus, Paula and M{\'e}ndez, Fernando J and Losada, Inigo J and Men{\'e}ndez, Melisa and Espejo, Antonio and P{\'e}rez, Jorge and Rueda, Ana and Guanche, Yanira.} (2014) A method for finding the optimal predictor indices for local wave climate conditions,  \textit{Ocean Dynamics.} \textit{64} (7),  {1025}--{1038}.

\bibitem[Charles et al. (2012)]{charles}
\textbf{Charles, Elodie and Idier, D{\'e}borah and Delecluse, Pascale and D{\'e}qu{\'e}, Michel and Le Cozannet, Gon{\'e}ri} (2012) Climate change impact on waves in the Bay of Biscay, France, \textit{Ocean Dynamics} \textit{62}(6),  {831}--{848}.

\bibitem[Chollet et al.(2015)]{keras}
\textbf{Chollet, Fran\c{c}ois and others.} (2015)  Keras, \url{https://keras.io}.

\bibitem[Diederik et al.(2017)]{adam}
\textbf{Diederik P. Kingma and Jimmy Ba} (2017) Adam: A Method for Stochastic Optimization,  \textit{arXiv.} \textit{3}.

\bibitem[Maraun et al.(2010)]{maraun}
\textbf{Goze M. and Remm E.} (2010)  Precipitation downscaling under climate change: Recent developments to bridge the gap between dynamical models and the end user,  \textit{WReviews of geophysics.} \textit{48}(3).

\bibitem[Hegermiller et al.(2017)]{heg}
\textbf{Hegermiller, CA and Antolinez, Jose AA and Rueda, A and Camus, Paula and Perez, Jorge and Erikson, Li H and Barnard, Patrick L and Mendez, Fernando J.} (2017) A multimodal wave spectrum--based approach for statistical downscaling of local wave climate, \textit{Journal of Physical Oceanography} \textit{47}(2),  {375}--{386}.

\bibitem[Hochreiter et al.(1997)]{lstm}
\textbf{Hochreiter, Sepp and Schmidhuber, J\"{u}rgen.} (1997)  Long Short-Term Memory,  \textit{Neural Comput.} \textit{9}(8),  {1735}--{1780}.

\bibitem[Ian Goodfellow et al. (2016)]{goodf}
\textbf{Ian Goodfellow and Yoshua Bengio and Aaron Courville.} (2016) Deep Learning,  \textit{MIT Press}.

\bibitem[Jiuxiang Gu et al.(2018)]{cnn}
\textbf{Jiuxiang Gu and Zhenhua Wang and Jason Kuen and Lianyang Ma and Amir Shahroudy and Bing Shuai and Ting Liu and Xingxing Wang and Li Wang and Gang Wang and Jianfei Cai and Tsuhan Chen.} (2018) Recent Advances in Convolutional Neural Networks,  \textit{Pattern Recognition}  \textit{77}), {354}--{377}.

\bibitem[Laugel et al.(2014)]{laugel}
\textbf{Laugel, Am{\'e}lie and Menendez, Melisa and Benoit, Michel and Mattarolo, Giovanni and M{\'e}ndez, Fernando.} (2014) Wave climate projections along the French coastline: dynamical versus statistical downscaling methods,  \textit{Ocean Modelling.} \textit{48},  {35}--{50}.

\bibitem[Michel et al.(2022)]{hcnn}
\textbf{Michel, M. and Obakrim, S. and Raillard, N. and Ailliot, P. and Monbet, V.} (2022) Deep learning for statistical downscaling of sea states,  \textit{Advances in Statistical Climatology, Meteorology and Oceanography.} \textit{8},  {83}--{95}.

\bibitem[Obakrim et al.(2022)]{paper}
\textbf{Obakrim, Said and Ailliot, Pierre and Monbet, Valerie and Raillard, Nicolas.} (2022) Statistical modeling of the space-time relation between wind and significant wave height, \textit{Earth and Space Science Open Archive.}

\bibitem[P{\'e}rez et al.(2014)]{estela}
\textbf{P{\'e}rez, Jorge and M{\'e}ndez, Fernando J and Men{\'e}ndez, Melisa and Losada, Inigo J} (2014) ESTELA: a method for evaluating the source and travel time of the wave energy reaching a local area,  \textit{Ocean Dynamics.} \textit{64} (8),  {1181}--{1191}.

\bibitem[Saha et al.(2010)]{cfsr}
\textbf{Saha, Suranjana and Moorthi, Shrinivas and Pan, Hua-Lu and Wu, Xingren and Wang, Jiande and Nadiga, Sudhir and Tripp, Patrick and Kistler, Robert and Woollen, John and Behringer, David and others} (2010) The NCEP climate forecast system reanalysis,  \textit{Bulletin of the American Meteorological Society.} \textit{91} (8),  {1015}--{1058}.

\bibitem[Wang et al.(2010)]{wang}
\textbf{Wang, Xiaolan L and Swail, Val R and Cox, Andrew.} (2010) Dynamical versus statistical downscaling methods for ocean wave heights,  \textit{International Journal of Climatology: A Journal of the Royal Meteorological Society} \textit{30}(3),  {317}--{332}.

\bibitem[Young, Ian R.(1999)]{waves}
\textbf{Young, Ian R} (1999) Wind generated ocean waves,  \textit{Elsevier.}

\end{thebibliography}
\end{document}